\documentclass{article}

\usepackage[nonatbib, preprint]{HighSchoolStyles/neurips_2024}

\usepackage[pdftex]{graphicx}
\usepackage[table,xcdraw]{xcolor}
\usepackage{colortbl}
\usepackage{caption}  

\usepackage[utf8]{inputenc}
\usepackage[T1]{fontenc}
\usepackage{hyperref}
\usepackage{url}
\usepackage{booktabs}       
\usepackage{amsfonts}
\usepackage{nicefrac}
\usepackage{microtype}
\usepackage{xcolor}

\hypersetup{
    colorlinks=true,
    urlcolor=blue
}

\title{ViMGuard: A Novel Multi-Modal System for Video Misinformation Guarding}

\author{
  Andrew Kan \\
  Weston High School \\
  \texttt{andrew.k@fusionresearch.co} \\
  \And
  Christopher Kan \\
  Noble and Greenough School \\
  \texttt{chris.k@fusionresearch.co} \\
  \And
  Zaid Nabulsi \\
  Advisor \\
  Google \\
  zaid.n@fusionreseaech.co
}

\begin{document}

\maketitle

\begin{abstract}
        The rise of social media and short-form video (SFV) has facilitated a breeding ground for misinformation. With the emergence of large language models, significant research has gone into curbing this misinformation problem with automatic false claim detection for text. Unfortunately, the automatic detection of misinformation in SFV is a more complex problem that remains largely unstudied. While text samples are monomodal (only containing words), SFVs comprise three different modalities: words, visuals, and non-linguistic audio. In this work, we introduce Video Masked Autoencoders for Misinformation Guarding (ViMGuard), the first deep-learning architecture capable of fact-checking an SFV through analysis of all three of its constituent modalities. ViMGuard leverages a dual-component system. First, Video and Audio Masked Autoencoders analyze the visual and non-linguistic audio elements of a video to discern its intention—specifically whether it intends to make an informative claim. If it is deemed that the SFV has informative intent, it is passed through our second component: a Retrieval Augmented Generation system that validates the factual accuracy of spoken words. In evaluation, ViMGuard outperformed three cutting-edge fact-checkers, thus setting a new standard for SFV fact-checking and marking a significant stride toward trustworthy news on social platforms. To promote further testing and iteration, VimGuard was deployed into a Chrome extension and all code was open-sourced on GitHub.
\end{abstract}

\section{Introduction}

The explosive growth of short-form video (SFV) platforms like TikTok and Instagram Reels has fundamentally reshaped how people consume and share content \cite{shortform}. These platforms cater to short attention spans, prioritizing quick, emotionally engaging content that spreads rapidly through algorithmic recommendation systems \cite{shortform}. While offering entertainment value, this format has also become a potent vector for misinformation. A concerning number of adults (48\%) now rely on SFVs as a primary news source, making them especially vulnerable to misleading narratives \cite{news}.

The rise of large language models (LLMs) has opened up possibilities for automating the fact-checking of online news. However, LLMs have well-documented limitations that make them less than ideal for the fact-checking of video content. Firstly, LLMs exhibit a tendency to hallucinate, generating plausible-sounding but factually incorrect information \cite{hallucination}. They are also trained on a static dataset and cannot access up-to-date information about the news or politics; this is a potentially disastrous shortcoming for a fact-checking algorithm. Secondly, LLMs are designed to analyze text, not video. Existing automated video fact-checking systems simply pass the text transcription of a video's audio (words that are spoken in the video) into an LLM or a system similar to an LLM (e.g. a text database) to determine whether any falsehoods are present \cite{misinformation_survey, hoes, maros}. This naive approach misses essential context found in the visual and non-linguistic audio portions of the video, leading to incorrect classifications. For example, a false statement that is made in a video does not warrant the video being taken down if it is made as part of a joke, in a song, or in a movie clip (i.e., if the false statement is not presented as fact).

In this work, we propose a novel approach to automated SFV fact-checking that addresses the limitations of traditional LLM-based methods. To address the problem of hallucinations and static information, we propose the use of an up-to-date database of news articles that the LLM can cross-reference (this is called Retrieval Augmented Generation) \cite{transformer, rag}. To address the problem of "missing context", we propose an initial Claim Detection task that will precede the LLM fact-checking of the video's transcript.

\begin{figure}[t]
    \centering
    \includegraphics[width=5in]{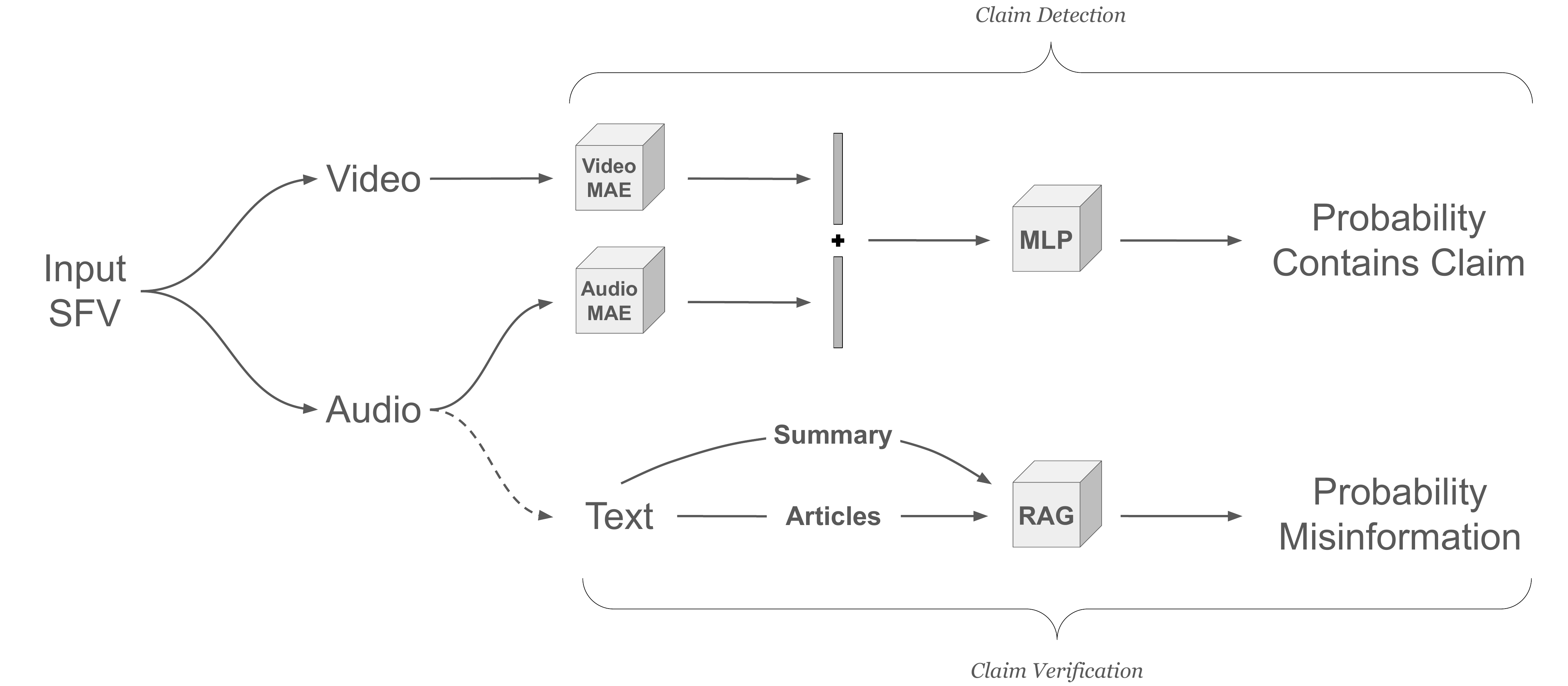}
    \caption{Overview of VimGuard's architecture. ViMGuard consists of two main components: Claim Detection (determining if a video has informative intent) and Claim Verification (determining if a video's claims are misinformative or not).}
\end{figure}

\section{Related Work}

Our work builds upon a growing body of research aimed at automating the fact-checking of short-form videos. Here's an overview of key areas and recent advancements:

\begin{itemize}

\item \textbf{Multimodal Fact-Checking}:  Researchers are developing methods to analyze the various modalities present in video content. This includes techniques for extracting and verifying text transcripts, assessing the visual components of a video (imagery, editing techniques), and analyzing audio elements for potential manipulation \cite{fact1, fact2, fact3, fact4, fact5}.

\item \textbf{Knowledge Base Integration}:  Efforts are underway to integrate external knowledge bases and fact-checking databases into automated systems. This helps systems compare claims made in videos against established facts  \cite{knowledge1, knowledge2, knowledge3}.

\item \textbf{LLMs for Fact-Checking}: Large language models have shown promise in generating explanations and summaries relevant to fact-checking, but struggle with reliability. Researchers are exploring fine-tuning techniques and ways to constrain LLM output to improve factual accuracy \cite{llm1, llm2, llm3}.

\item \textbf{Self-Supervised and Masked Autoencoders}:   The use of self-supervised learning (SSL) is gaining attention in the field of fact-checking. SSL allows models to learn from unlabeled data, potentially making them more robust. Masked Autoencoders, a type of generative SSL, focuses on teaching models to learn the inherent characteristics of objects in the training images \cite{mae1, mae2, mae3, dls, merlin}.

\end{itemize}

\section{Methods}

The goal of our model is to determine if an SFV contains misinformation. We frame our problem as two consecutive binary classification tasks, with the second binary classification gated on the first (an overview of our architecture can be found in Figure 1). The first stage, Claim Detection, determines whether or not the SFV has informative intent and needs to be fact-checked (i.e., whether or not the SFV contains an informative claim). If not, the SFV is classified as harmless. If so, the video is passed through our second stage component, Claim Verification, which fact-checks spoken words in the SFV with a Retrieval Augmented Generation (RAG) LLM system. If the claims made in the SFV are true, the SFV is classified as harmless. If the claims are false, the SFV is classified as misinformative.

As was briefly mentioned in Section 1, the reasoning for using this two-part formulation instead of a simple binary (misinformation or not misinformation) classification task is due to the limited ability of LLMs to understand and analyze video content. Although LLMs (especially those assisted by RAG) can successfully discern true and false statements in isolation, they struggle to do so within the complicated context of a video. For example, the statement "the London Bridge is falling down" would be flagged by an LLM as misinformation, even if this "claim" was being made in the context of a harmless nursery rhyme video. This is typically not a problem on news or discussion platforms, in which nearly every post is making an informative claim that requires fact-checking. On SFV platforms, however, the majority of videos are purely for entertainment, thus leading to many of these false positive classifications (harmless videos being classified as misinformative). To circumvent this issue, our Claim Detection task analyzes the visual and non-linguistic audio portions of the video to weed out videos that likely do not present any information. This task can be thought of as watching a video in a foreign language: you are unable to recognize any words being spoken, but you are still able to distinguish between a nursery rhyme or dance video (which likely does not contain an informative claim) from a podcast or news clip (which likely does contain an informative claim).

\begin{figure}[t]
    \centering
    \includegraphics[width=3in]{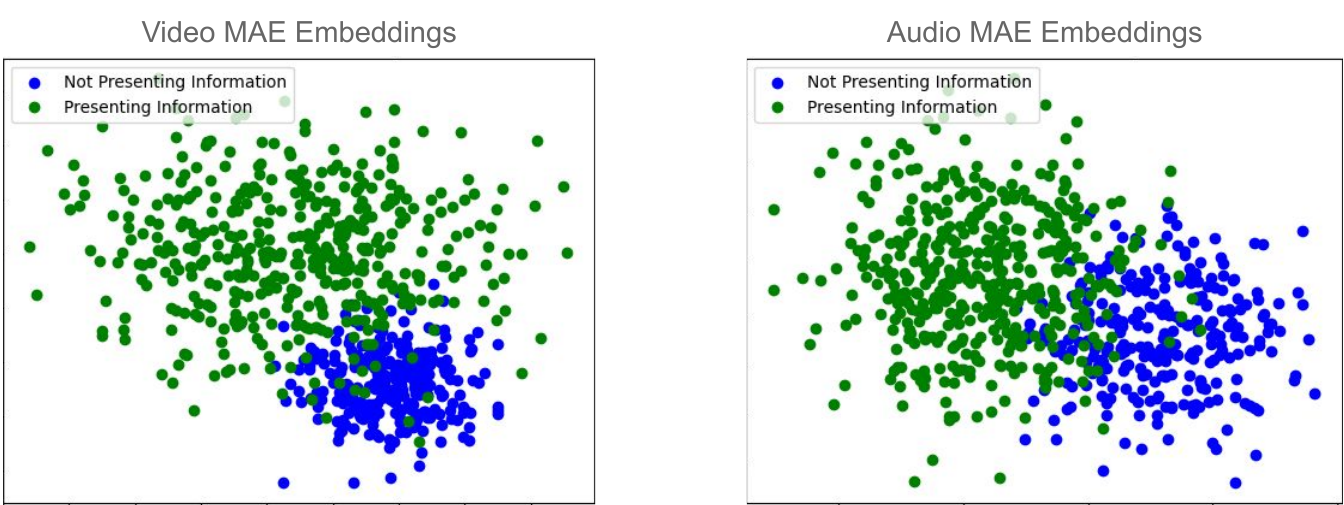}
    \caption{A visualization of the embedding outputs of Claim Detection pretraining. Each point is an embedding downsampled to 2 dimensions using principal component analysis.}
\end{figure}

\subsection{Claim Detection with Masked Autoencoders}
Because of the dearth of labeled data, we train the Claim Detection model in two stages: a pretraining stage and a supervised fine-tuning stage. Both stages are detailed below.

\subsubsection{Pretraining}

The goal of our pretraining task is to develop a generalist model familiar with the foundational features of an SFV’s audio and visual spaces. Pretraining is critical here because there is very little labeled data pertaining to informative and non-informative SFVs; pretraining leverages large amounts of unlabeled video data so that less labeled data is needed. 

Our pretraining task utilizes a cutting-edge algorithm called Masked Autoencoders (MAE) \cite{mae1}. MAEs divide images into patches and mask approximately 75\% of pixels before converting the data into embeddings. However, video content introduces a temporal dimension to the spatial layout, demanding a three-dimensional approach \cite{mae2, tb, mrngan}. Video MAEs tackle this issue by dividing videos into three-dimensional cubes of width, height, and time, allowing the encoder to effectively learn from the footage.

We adopt a similar approach for audio utilizing the Audio MAE algorithm \cite{mae3}. Audio MAE converts the audio to a spectrogram, an image, where the time domain is on the horizontal axis and the frequency is on the vertical axis. The spectrogram is then masked like a regular image, allowing the model to capture the underlying structure of the audio signal.

Separate MAE models are trained for the visual and audio components (with different masking ratios, masking shapes, downsampling, etc.). Each takes an SFV as input and outputs an embedding representation of their respective modality \cite{mae2, mae3, latent, lung_cancer}. The embeddings are from the Video and Audio MAEs are concatenated to form a final embedding output of the pertaining task. Our MAEs were trained on a dataset of around 32 million TikTok videos scraped in 2020.

See Figure 2 for a downsampled representation of the embedding outputs of our two MAE models. It is clear from these graphs that our MAE models have significant discriminative ability.
 
\subsubsection{Supervised Finetuning}

Once the generalist model (the MAE from pretraining) has become familiarized with social media video content, the finetuning stage hones this model to identify videos as either presenting an informative claim or not \cite{hear, poverty}. This is done by appending a feed-forward neural network to the generalist model that will classify its embedding outputs.

In training, labeled examples of SFVs (presenting claims, and not presenting claims) are fed through the generalist model and converted into embeddings. The feed-forward neural network learns to classify these embeddings as either presenting a claim or not presenting a claim.

To collect labeled examples of SFVs that are presenting claims and not presenting claims, we scraped around 5000 videos off of TikTok using the TikTok API. We scraped videos by their hashtags and then manually ensured each video that was scraped fell into its intended category. The following hashtags were used for the “contains a claim” label: \#podcast, \#news, \#politics, \#election, \#health, \#fitness, \#nutrition, \#science, \#history, \#technology, \#investing, \#finance. And the following were used for the “does not contain a claim” label: \#dance, \#music, \#challenge, \#memes, \#prank, \#skit, \#standup, \#gaming, \#movies, \#art, \#fashion, \#beauty, \#diy, \#cooking, \#travel, \#adventure, \#pets.

\subsection{Claim Verification with Retrieval Augment Generation}

The first step of Claim Verification involves transcribing the audio component of the SFV to text. This task is accomplished using Whisper, a model developed by OpenAI capable of transcribing audio clips. The transcribed audio is then summarized by OpenAI’s GPT-4.

Next, the summarized transcription of the SFV is analyzed with Elasticsearch, an algorithm capable of managing and sifting through vast quantities of text data. Elasticsearch examines the summary for key terms, which are then queried into a database of text articles. The database returns articles that pertain to the SFV. (The database we used for Elasticsearch was manually collated from the following sources: Wikipedia, Google Claim Review, and Bing News. Access to these sources allows our final Claim Verification model to speak on current events, politics, etc.)

 The relevant articles obtained through Elasticsearch and the initial text summary of the SFV are finally fed back into GPT-4, which determines whether the video's claims are true or false---this process of querying an LLM and giving it external data is called Retrieval Augmented Generation. By analyzing claims with its pretrained dataset and through comparison with up-to-date news articles, GPT-4 can make a more informed judgment.

\begin{table}[htbp]
\renewcommand{\arraystretch}{1.25}
\centering
\captionsetup{skip=10pt}
\caption{Performance comparison of ViMGuard and other text fact-checkers}
\label{tab:performance_comparison}
\begin{tabular}{lccc}
\toprule
\textbf{Model}                & \textbf{AUROC} & \textbf{F1 Score} & \textbf{\# API Calls} \\
\midrule
ViMGuard                       & 86.3\%          & 89.3\%             & 283                   \\
UTA's ClaimBuster              & 79.8\%          & 81.3\%             & 734                   \\
Hoes, et al. (2023)               & 63.1\%          & 59.8\%             & 734                   \\
Google ClaimReview             & N/A             & 58.9\%             & 734                   \\
\bottomrule
\end{tabular}
\end{table}

\section{Evaluation and Discussion}

ViMGuard is the first robust, end-to-end fact-checker designed for SFV content (to our knowledge), so in evaluation, we benchmarked our model's performance against three cutting-edge text fact-checkers (See Table 1 for results). The fact-checking of an SFV with a text fact-checker is accomplished by simply passing the text transcription of the SFV's audio into the fact-checker. We benchmarked against two database-reliant text fact-checkers (University of Texas at Austin's ClaimBuster \cite{claimbuster} and Google ClaimReview) as well as a fact-checking implementation of an LLM from 2023 \cite{hoes}.

The dataset used for evaluation was a handmade dataset comprised of SFVs scraped off of TikTok, Instagram Reels, and Youtube Shorts, as well as some homemade SFVs (videos we recorded on a phone). Our final dataset contained around 700 SFVs, each of which was manually labeled as either "misinformative" or "not misinformative." Our goal was to create a sizeable sample of videos representative of what a typical feed looks like on an SFV platform.

In short, ViMGuard outperformed all other models in the two proxies for accuracy that were measured---area under the receiver operating characteristic curve (AUROC) and F1 score. ViMGuard's outperformance of the LLM fact-checker (by 23\% in AUROC and by 30\% in F1 score) demonstrates that ViMGuard is not only effective because of its use of GPT-4 and that our novel Claim Detection task and use of RAG has a substantial impact on final performance. The next best model that was benchmarked against, UTA's ClaimBuster, had a significantly larger, professionally collated news database to cross-reference and thus achieved a respectable accuracy despite its lack of LLM use or a Claim Detection task. It can be expected that ViMGuard's performance will increase considerably if given access to a news database like ClaimBuster's. 

ViMGuard also outperformed all other models in efficiency (representatively measured by the number of API calls made to either an LLM or article database) due to our Claim Detection task. The difference in speed between a feed-forward neural network classification and a database or LLM call is significant, especially when considering that an SFV fact-checker like ViMGuard could be used to vet millions of videos a day. Gating our pipeline on a simpler Claim Detection task thus increases both accuracy and efficiency. ViMGuard's unprecedented performance---despite only having access to a small number of open-source news articles---demonstrates the soundness of our architecture and its potential for scalability.

\section{Supplementary Artifacts}
    To promote further testing and iteration, ViMGuard was deployed into a \href{https://drive.google.com/file/d/193CdUUAF3uJ5Xpv8ZJwN61RvHXk6PBeH/view?usp=sharing}{Chrome extension} and all code was open-sourced on \href{https://github.com/ckant0/ViMGuard}{GitHub}. View ViMGuard's \href{https://docs.google.com/presentation/d/12ua_hPpbpojCCF9IcGcqz-yxTPeWHP3GSEvFald-COQ/edit?usp=sharing}{digital poster} for more information.

\newpage

\begin{ack}

We would like to thank Zaid Nabulsi for his mentorship throughout this project. Zaid advised us in the training and evaluation of our models---namely, he helped us tune hyperparameters (number of layers, learning rate, etc.) and advised us on how to assess ViMGuard most comprehensively (what metrics to test). Zaid also gave us access to the computing resources at his research institute, which included 4 Nvidia V100 GPUs.

\end{ack}

{\small
\bibliographystyle{ieee}
\bibliography{egbib}
}

\end{document}